\documentclass[conference]{IEEEtran}
\IEEEoverridecommandlockouts
\usepackage{cite}
\usepackage{amsmath,amssymb,amsfonts}
\usepackage{algorithmic}
\usepackage{graphicx}
\usepackage{textcomp}
\usepackage{xcolor}
\usepackage{subfigure}
\def\BibTeX{{\rm B\kern-.05em{\sc i\kern-.025em b}\kern-.08em
    T\kern-.1667em\lower.7ex\hbox{E}\kern-.125emX}}
\begin{document}

\title{ConViViT - A Deep Neural Network Combining Convolutions and Factorized Self-Attention for Human Activity Recognition\\}

\author{
\IEEEauthorblockN{DOKKAR Rachid Reda}
\IEEEauthorblockA{\textit{Efrei Research Lab} \\
\textit{Paris Pantheon-Assas University}\\
Paris, France \\
hr\_dokkar@esi.dz}
\and
\IEEEauthorblockN{Faten Chaieb}
\IEEEauthorblockA{\textit{Efrei Research Lab} \\
\textit{Paris Panthéon-Assas University}\\
Paris, France \\
faten.chakchouk@efrei.fr}

\and
\IEEEauthorblockN{Hassen Drira}
\IEEEauthorblockA{\textit{ICube UMR 7357, CNRS} \\
\textit{Université de Strasbourg}\\
Strasbourg, France \\
hdrira@unistra.fr}

\and
\IEEEauthorblockN{Arezki Aberkane}
\IEEEauthorblockA{\textit{Caplogy SAS} 
Vélizy-Villacoubay, France \\
a.aberkane@caplogy.com}
}

\maketitle

\begin{abstract}
The Transformer architecture has gained significant popularity in computer vision tasks due to its capacity to generalize and capture long-range dependencies. This characteristic makes it well-suited for generating spatiotemporal tokens from videos. On the other hand, convolutions serve as the fundamental backbone for processing images and videos, as they efficiently aggregate information within small local neighborhoods to create spatial tokens that describe the spatial dimension of a video. While both CNN-based architectures and pure transformer architectures are extensively studied and utilized by researchers, the effective combination of these two backbones has not received comparable attention in the field of activity recognition. In this research, we propose a novel approach that leverages the strengths of both CNNs and Transformers in an hybrid architecture for performing activity recognition using RGB videos. Specifically, we suggest employing a CNN network to enhance the video representation by generating a 128-channel video that effectively separates the human performing the activity from the background. Subsequently, the output of the CNN module is fed into a transformer to extract spatiotemporal tokens, which are then used for classification purposes. Our architecture has achieved new SOTA results with 90.05 \%, 99.6\%, and 95.09\% on HMDB51, UCF101, and ETRI-Activity3D respectively.
\end{abstract}

\begin{IEEEkeywords}
Activity Recognition, Transformer, CNN

\end{IEEEkeywords}

\section{Introduction}
Activity recognition can be defined as allowing the machine to recognize/detect the activity based on information received from different sensors. These sensors can be cameras, wearable sensors, and sensors attached to objects of daily use or deployed in the environment. In this work, we are interested in activity Recognition using RGB videos. RGB videos are a complex type of data, that contain complex spatiotemporal dependency. 

To extract spatial features, we utilized a CNN module inspired by \cite{uniformer}. The primary objective of this CNN module is to enhance the representation of the video by extracting spatial tokens.

The advantages of applying CNNs to video data have been extensively discussed in \cite{uniformer}. It has been affirmed that by applying small filters to localized neighborhoods of pixels, CNNs are capable of extracting fine-grained spatial tokens. Furthermore, it has been demonstrated that CNNs outperform self-attention mechanisms in terms of spatial token extraction. Moreover, by employing CNNs as a first step, the subsequent transformer module is able to perform self-attention on a reduced spatial set of tokens and extract temporal dependencies. 

To extract temporal features, we have used a video transformer architecture inspired by \cite{video-vit}. Within this framework, we have investigated two distinct architectures, each utilizing a different type of self-attention: Factorised Dot-Product and Factorised Self-attention. Both of these attention mechanisms apply self-attention to both spatial and temporal axes. Factorized Self-attention first applies spatial attention, followed by temporal attention on the output of the spatial attention. In contrast, the Dot-Product attention applies spatial and temporal attention to the input and subsequently fuses the results. We have proved through our experiments that factorized self-attention yields better results compared to other approaches.

The proposed architecture has achieved state-of-the-art performance on three benchmark datasets. We conducted tests on HMDB51 \cite{dataset-hmdb51}, UCF101 \cite{dataset-ucf-101} and ETRI-Activity3D \cite{etri-3d} and obtained state-of-the-art results on HMDB51, UCF101, and ETRI-Activity3D with 90.05 \%, 99.6\%, and 95.09\% respectively.

\section{ Related Works}
The existing literature on activity recognition and computer vision can be broadly categorized into three main groups: (1) CNN-based approaches, (2) Transformer-based approaches and (3) Hybrid architectures combining CNNs and Transformers.

\subsection{CNN-based approaches}
3D Convolutional Neural Networks (CNNs) have traditionally been the primary choice for visual data processing, encompassing various types of visual data such as images and videos. Consequently, they have held a dominant position in the field of computer vision for a considerable time. However, with the adaptation of attention mechanisms and the transformer architecture from Natural Language Processing (NLP) to Computer Vision (CV), the landscape has witnessed significant changes \cite{vision-transformer}.

Previous studies have attempted to address the challenges of activity recognition using purely 3D and 2D Convolutional Neural Networks \cite{rw-cnn-1, rw-cnn-2}. However, optimizing the results and achieving satisfactory performance with a pure CNN architecture for activity recognition has proven to be challenging, primarily due to the high computational demands associated with these architectures.

To overcome these challenges, the I3D approach \cite{i3D-cnn} introduced the concept of inflating pre-trained 2D convolution kernels, which allowed for better optimization of the network. In addition, other prior works focused on factorizing 3D convolution kernels in various dimensions to reduce computational complexity \cite{rw-cnn-3, rw-cnn-3-1, rw-cnn-4} .

More recent studies have proposed techniques to enhance the temporal modeling ability of 2D CNNs \cite{rw-cnn-5, rw-cnn-6}. However, due to the inherent nature of CNNs, which aggregate information within a small window of the neighborhood, these approaches did not achieve significant improvements in performance.

Taken together, these prior works have explored different strategies to address the challenges of activity recognition using CNN architectures. While attempts have been made to optimize and enhance the performance of pure CNNs, limitations related to computational requirements and the inherent nature of CNNs' spatial aggregation persist.

\subsection{Transformer-based approaches}

Since the introduction of the vision transformer \cite{vision-transformer}, numerous studies have embraced the transformer architecture for computer vision tasks. These works have consistently surpassed the results achieved by CNNs. This is due to the transformers' ability to capture long-range dependencies and effectively attend to important regions of the input through self-attention mechanisms.

Several notable works have contributed to the adoption and advancement of the transformer architecture in computer vision. These include works such as \cite{rw-transformer-1}, \cite{rw-transformer-2}, \cite{rw-transformer-3}, \cite{rw-transformer-4}, which propose various variants for spatiotemporal learning in video analysis. These variants aim to harness the power of transformers in capturing both spatial and temporal information for more comprehensive video understanding. 

Video Vision Transformers (Timesformer \cite{pmlr-v139-bertasius21a} and ViViT \cite{video-vit}) are among the early Transformers approaches for action recognition. They introduce innovative embedding schemes and adaptations to ViT \cite{rw-transformer-3} and other related Transformers for modeling video clips. 

In \cite{pmlr-v139-bertasius21a}, the authors propose a tokenization scheme called uniform frame sampling based on a randomly selected frames from a video.
Along similar lines, ViViT \cite{video-vit} introduced Tubelets Embedding to effectively preserve contextual time data within videos and handles 3D volumes instead of frames. Four different variants were proposed based on the attention technique: Spatiotemporal attention, Factorized Encoder, Factorized self-attention, and Factorized dot-product attention.

Simultaneously, other research efforts have focused on mitigating the computational cost associated with transformer architectures while still achieving impressive results. An example of such work is the Swin Transformer \cite{swin-transformer}, which presents an innovative architecture designed to strike a balance between computational efficiency and powerful performance.

Collectively, these works have significantly propelled the adoption of transformer architectures in computer vision. By capitalizing on their ability to capture long-range dependencies and leverage self-attention mechanisms, these architectures have demonstrated remarkable capabilities in various visual tasks.

The self-attention mechanism is considered inefficient when it comes to encoding low-level features. To address this limitation, the Swin Transformer approach introduces a solution by applying attention within a local 3D window. This localized attention mechanism allows for more efficient encoding of low-level features.

\subsection{Hybrid approaches}
In recent research, efforts have been made to incorporate convolutional neural networks into the transformer architecture for image recognition tasks. However, these approaches have not adequately addressed the spatiotemporal aspect of videos. Recognizing this limitation, Uniformer \cite{uniformer} presented an architecture specifically tailored for video understanding, utilizing a concise transformer format based on 3D convolutions that unifies convolutions and transformers. By integrating convolutional operations into the transformer architecture, Uniformer aims to combine the strengths of both convolutional neural networks and transformers, resulting in an improved framework for feature encoding. 

In this work, we adopt 3D convolutions to address the inefficiency of self-attention in encoding low-level features. In fact, 3D convolutions enhance the capability of our model to capture both spatial and temporal information effectively.

\section{Proposed Method}
\begin{figure*}[!ht]
\centering
\includegraphics*[scale=0.18]{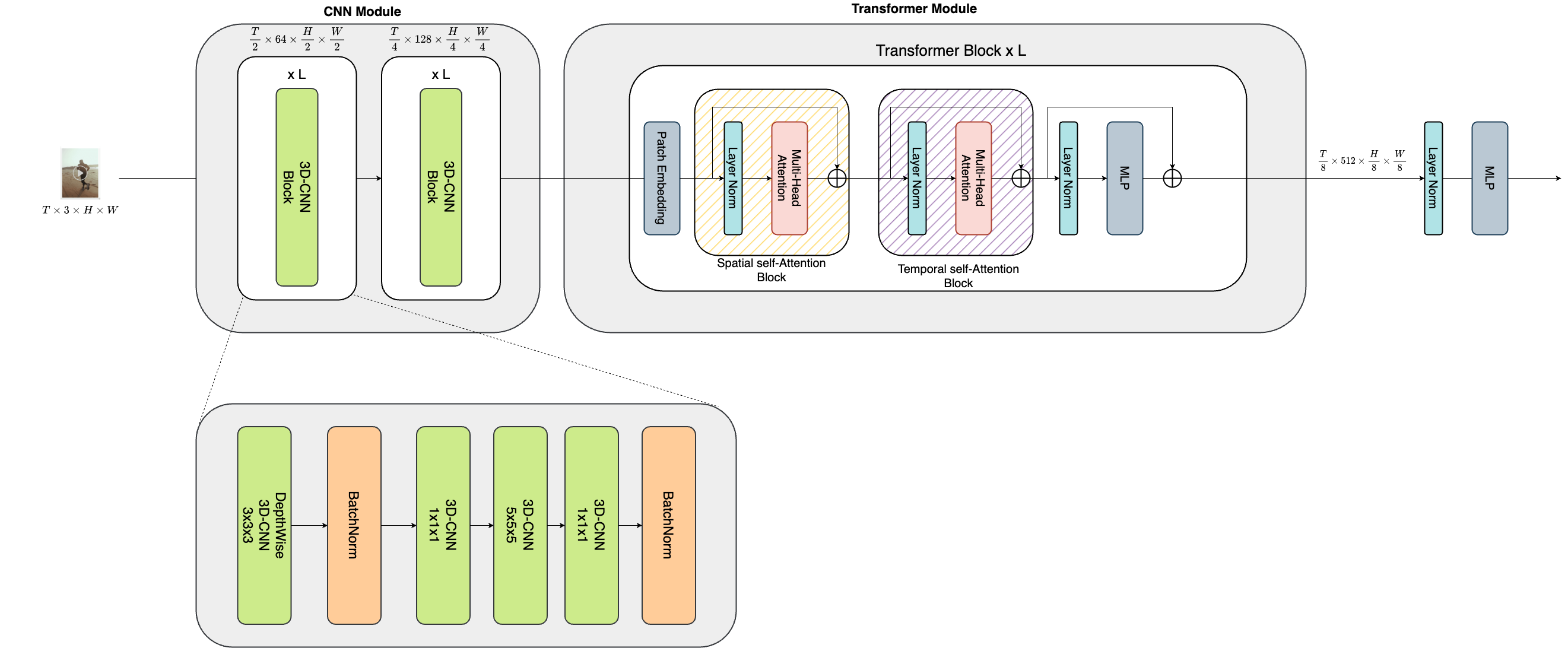}
\caption{Overall proposed architecture for Human Activity Recognition}
\label{fig:my-arch}
\end{figure*}

\subsection{Architecture overview}

The proposed architecture consists of two main modules, as depicted in Figure \ref{fig:my-arch}: a CNN module to extract spatial features followed by a transformer module. 

The CNN module plays a crucial role in capturing spatial information from the input data. It applies convolutional operations to extract relevant features that describe the spatial characteristics of the input images. The primary objective of this CNN module is to enhance the representation of the video by extracting spatial tokens. Its output is later fed to the transformer module.

The transformer module is the factorized self-attention transformer proposed in \cite{video-vit}. It takes advantage of its self-attention mechanism to capture long-range dependencies and model the interactions between spatial features. This module leverages the encoded spatial information from the CNN Module to extract spatiotemporal features that are vital for accurate action classification.

Inspired by the contributions of these two works, we propose an hybrid approach that combines the strengths of CNNs for extracting spatial cues and the transformer architecture for extracting spatiotemporal tokens. Our CNN module aims to enhance the video representation by transforming it from a three-channel video to a 128-channel video. Subsequently, the transformer takes the output of the CNN module and applies a Patch Embedding similarly to \cite{uniformer}. Then a factorized self-attention is applied generating a rich spatiotemporal representation that will be used by the classification head.

\subsection{Patch Embedding}

The main purpose of Patch Embedding is to provide the order information to the transformer by slicing the input into $16\times16$ patches. Our proposed patch embedding block is inspired by the design and implementation of Uniformer's Dynamic Position Embedding (DPE) architecture \cite{uniformer} since the use of DPE improves the state-of-the-art results by $0.5\%$ and $1.7\%$ on ImageNet and Kinetics-400, respectively. This shows that by encoding the position information, DPE can maintain the spatiotemporal order, thus contributing to better learning of the spatiotemporal representation \cite{uniformer}.

The main contribution here is the use of a CNN layer with $0$ padding to create a more adequate representation. The fact that the transformer does not take video as input makes the application of a CNN layer more advantageous because this layer allows to manipulate and adjust how we introduce the order information to obtain better results.

\subsection{CNN Block}

The proposed CNN block aims to extract spatial information and offers a compact spatial representation to be fed to the transformer block. It is based on 3D-CNNs and Depth-Wise CNNs architectures (see figure \ref{fig:my-arch}) as follows:

\begin{itemize}
\item \textbf{DW 3D-CNN $3\times3\times3$}: the depth-wise 3D-CNN aims to extract the spatial features of each 3D neighborhood ($3\times3\times3$). It consists in applying a single convolutional filter for each input channel. This allows to better extract spatial features. 
\item \textbf{3D-CNN $1\times1\times1$}: the 3D-CNN ($1\times1\times1$) aims to reduce the dimension of the input before applying the ($5\times5\times5$) filter to save computation time.
\item \textbf{3D-CNN $5\times5\times5$}: The application of a 3D filter of size 5 (larger than 3) allows to have a more global representation of the spatial neighborhood.
\item \textbf{3D-CNN $1\times1\times1$}: the 3D-CNN ($1\times1\times1$) is intended to increase the size of the input to give more information to the following steps.
\end{itemize}

\subsection{Spatiotemporal Transformer Module}

The transformer block is the most important part of the proposed architecture. It takes the output of the CNN Module (spatial representation) in order to create a spatiotemporal representation. The application of attention allows any transformer to focus on the most important parts of the input as well as to understand the long sequences which allows extracting the temporal dependencies from a spatial sequence.  

To extract temporal features, two types of self-attention could be chosen, the factorized dot product and the factorized self-attention proposed in \cite{video-vit}. Both apply attention to the spatial and temporal axes.

\paragraph{\textbf{Factorised self-attention }} 
It consists in applying attention on the spatial axis of the input followed by temporal attention.
The application of attention on the spatial axis first allows us to take into consideration the dependencies between the spatial tokens and to deduce the most important parts that characterize this axis, thus preparing the input for the next operation of self-attention. The result of the self-attention on the spatial axis will be the input for the self-attention on the time axis that has the goal of the extraction of the spatiotemporal features. 

\paragraph{\textbf{Factorized Dot-Product attention }}
Factorized self-attention is a way of applying self-attention. The idea is to apply attention on the spatial axis of the input X and then have an output Y that we will apply attention on its temporal axis.

The application of attention on the spatial axis first allows to take into consideration the dependencies between the spatial tokens and to deduce the most important parts that characterize this axis, thus preparing the input for the next operation of self-attention. The result of the self-attention on the spatial axis will be the input for the self-attention on the time axis that has for the goal of the extraction of the spatiotemporal features.

Although both variants are interesting, the factorized attention seems to be more suitable for our architecture (non-video inputs). 

\section{Experiments}
\label{Sec:Experiments}

\subsection{Datasets}

\begin{itemize}
\item \textbf{ETRI-3D} \cite{etri-3d}: ETRI-3D is the first large-scale RGB-D dataset of the daily activity of the elderly ($112\;620$ samples). ETRI-3D is collected by Kinect v2 sensors and consists of three synchronized data modalities: RGB video, depth maps, and skeleton sequences. To film the visual data, 50 elderly subjects are recruited. The elderly subjects are in a wide age range from 64 to 88 years old, which leads to a realistic intra-class variation of actions. In addition, they acquired a dataset for 50 young people in their twenties in the same manner as the elderly. %Finally, 112,620 3D datasets were obtained.

\item \textbf{UCF101} \cite{dataset-ucf-101}: UCF101 is an action recognition dataset of realistic action videos collected from Youtube with 101 action categories. With 13320 videos of 101 action categories, UCF 101 offers a wide variety of actions with the presence of large variations in camera movement, object appearance and pose, object scale, viewpoint, cluttered background, lighting conditions, etc.

\item \textbf{HMDB51} \cite{dataset-hmdb51}: The HMDB51 dataset is a large collection of realistic videos from a variety of sources, including movies and web videos. The dataset consists of 6,849 video clips from 51 action categories (such as "jump" and "laugh"), with each category containing at least 101 clips. The original evaluation scheme uses three different training/test divisions. Within each division, each action class has 70 clips for training and 30 clips for testing.

\end{itemize}

\subsection{Visualization of ConViViT shallow and Deep Layers outputs}

\begin{figure*}[!ht]
\centering
\subfigure[Output of the first 3D CNN Block]{\includegraphics[scale=.3]{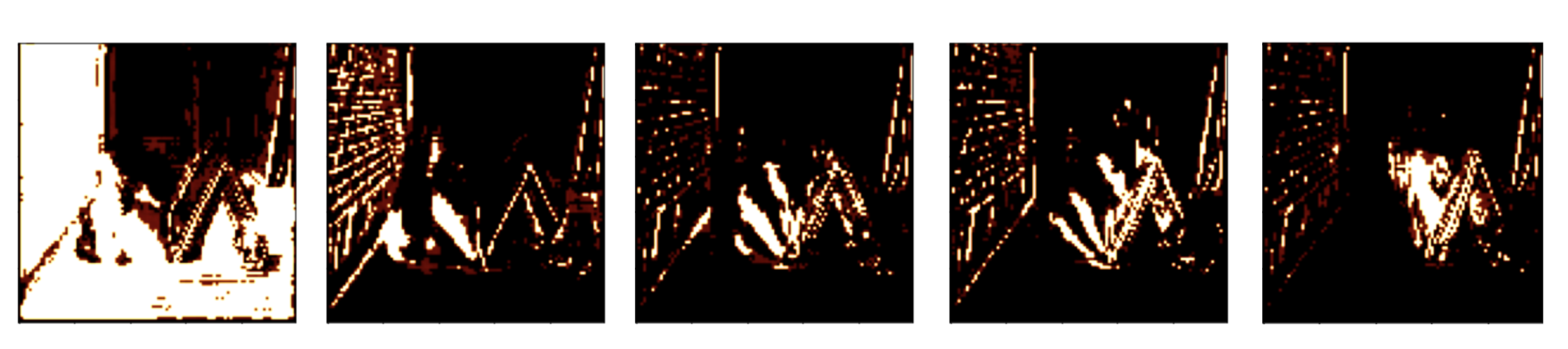}
\label{CNN1}}
\subfigure[Output of the CNN Module]
{ 
\includegraphics[scale=.3]{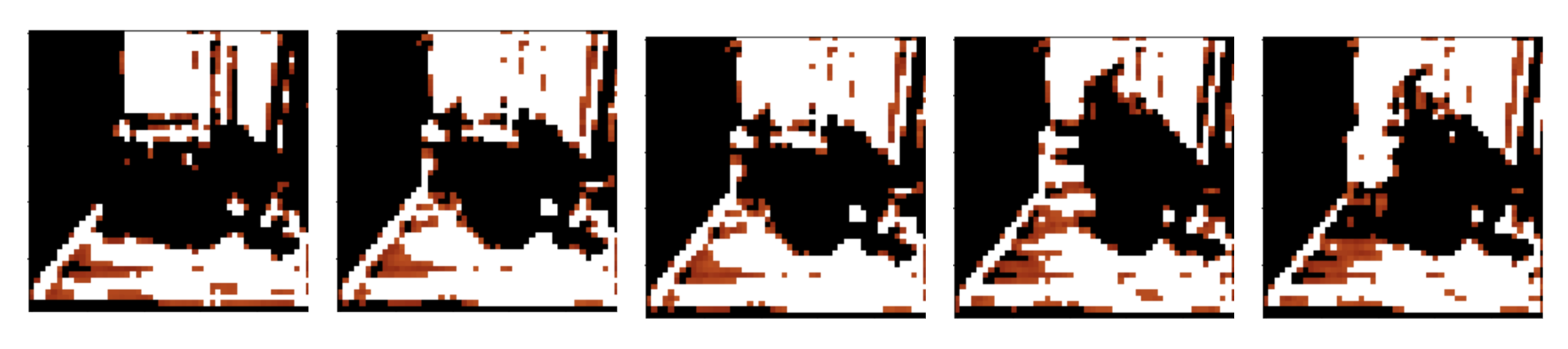}
\label{CNN2}}
%\hfill
\subfigure[Attention maps of ConViViT]
{
        % \centering
         \includegraphics[scale=.3]{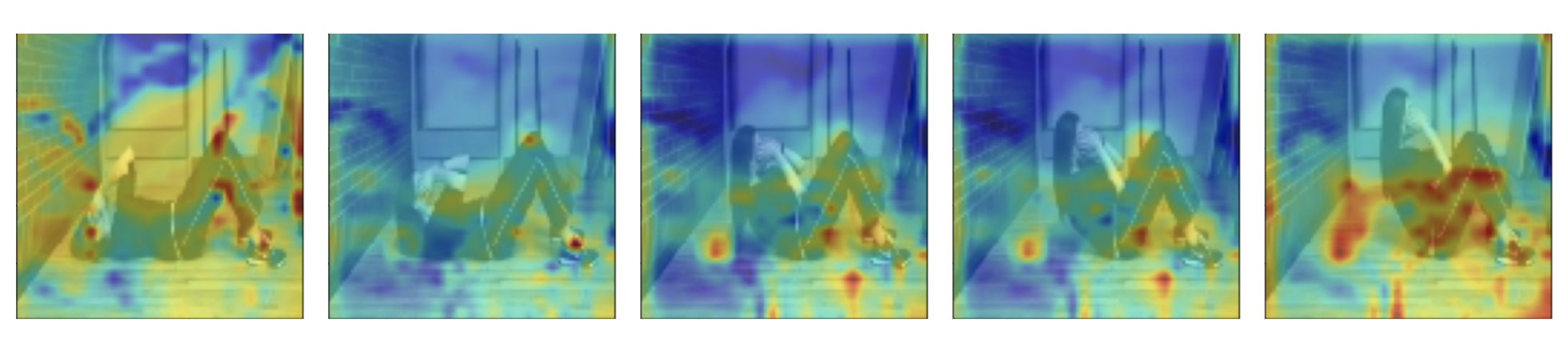}
         \label{fig:heatmap}
    }
\caption{Visualization of spatial and spatiotemporal Transformer modules outputs on a clip from HMDB51 dataset}
\label{fig:model-viz}
\end{figure*}

Figure \ref{fig:model-viz} is a visualization of the effect of our spatial and spatiotemporal block in the shallow and deep layers. 

We observe that the output of the first CNN block allows to localize the person (see figure \ref{CNN1}). Then the output of the next 3D-CNN block which is the output of the spatial module gives more importance to the person who makes the action (see figure \ref{CNN2}). Figure \ref{fig:heatmap} shows five attention maps of ConViViT computed for a sequence of five images showing an action from HMDB51 dataset. 
Each attention map refers to the visualizations of the attention weights computed between each patch in the image.  The last attention map which refers to the last frame of the action sequence shows that our model has succeeded to capture the entire trajectory of the action (red zones).

\subsection{Ablation study}

In this section, we will investigate the influence of different modules of the proposed architecture on the overall performance. All experiments in this study are conducted on the HMDB51 dataset.

Mainly we focus on the usefulness of using a CNN block to extract spatial features as well as the usefulness of factorized attention.

As a reminder, the factorized attention block is inspired by the Vivit \cite{video-vit} architecture and the CNN block by Uniformer \cite{uniformer}. In order to study the impact of CNN and factorized attention, we compare the variants of the proposed architecture with Uniformer and Vivit. So, the Vivit architecture is based only on the transformer which applies the factorized attention directly to the input. The Uniformer consists of a hybrid architecture based on the general formula of attention introduced in the first vision transformer \cite{vision-transformer}.

\subsubsection{Factorized Attention vs Factorized Dot Product Attention}
\label{sec:ablation-FA-vsFADP}

We compare the two variants of self-attention: The factorized self-attention used in our architecture and the factorized dot product. 

As illustrated in Table \ref{tab:Factorized vs Dot Product} (first two rows), the factorized attention outperforms the factorized dot-product architecture.

\subsubsection{Impact of the spatial block}
\label{sec:ablation-block-s}
~\\
To illustrate the importance of the proposed spatial block, we compare our results with those obtained by Vivit \cite{video-vit}.
The table \ref{tab:Factorized vs Dot Product} (first and third rows) shows that the proposed architecture exceeds Vivit by $9.87\%$. We opted for a hybrid architecture of CNN and transform it to prepare a spatial representation before applying any type of attention. The result of our test with dot-product attention supports our hypothesis regarding the use of CNN before transformers in computer vision.

In order to study the impact of the number of the required CNN blocks in the CNN Module, we compare the results obtained by our architecture based on a single CNN Block with two CNN Blocks.
We notice that using two CNN Blocks yields to $90.05\%$ accuracy versus $64.47\%$ using one CNN block (see Table \ref{tab:Factorized vs Dot Product} - first and fourth rows).

\subsubsection{Spatio-temporal block}
\label{sec:ablation-block-st}

To prove the value of our spatio-temporal block we can compare our results with the results of Uniformer \cite{uniformer} since they opted for a hybrid architecture too but they used the general formula of attention introduced in the first vision transformer \cite{vision-transformer}. This experiment reveals an improvement of accuracy by $4.69\%$ when comparing our architecture ($90.05\%$) to first vision transformer \cite{vision-transformer} ($85.36\%$) (see Table \ref{tab:Factorized vs Dot Product} - First and fifth rows).
 
Actually, applying normal attention to the output of CNN gives good results but our choice to apply factorized attention is better because applying attention on an axis allows the model to see the dependencies between the elements on that axis. Thus, by applying attention to the spatial and then temporal axis we obtained results that exceeded that of Uniformer.

\begin{table}[!h]
\caption{Ablation study on HMDB51 dataset}
\label{tab:Factorized vs Dot Product}
\centering
\begin{tabular}{|l|c|}
\hline
\textbf{Model} & \textbf{Accuracy} \\ \hline
\multicolumn{1}{|l|}{\textbf{Ours - Factorized Self-attention variant}} & \textbf{90.05\% }\\ \hline
\multicolumn{1}{|l|}{Ours - Factorized Dot Product self-attention variant } & 81.74\% \\ \hline
Vivit - Factorized Self-attention \cite{video-vit} & 80.18\% \\ \hline
Ours -  Self-attention Variant (One CNN Block)  & 64.47\% \\ \hline
UniFormer \cite{uniformer} & 85.36\% \\ \hline
\end{tabular}%

\end{table}

\subsection{Comparison with state-of-the-art}

\begin{table}[!h]
\caption{Accuracy comparison on HMDB51}
\label{tab:comp-us-vs-others-hmdb51}
\centering
%\resizebox{0.73\width}{!}{
\begin{tabular}{|l|c|c|}
\hline
\textbf{Architecture} & \textbf{Accuracy} & \textbf{Pre Training} \\ \hline
DEEP-HAL with ODF+SDF \cite{hmdb-1} & 87.56\% & Yes \\ \hline
TO+MaxExp+IDT \cite{hmdb-2} & 87.21\% & Yes \\ \hline
SCK \cite{hmdb-3} & 86.11\% & Yes \\ \hline
SO+MaxExp+IDT \cite{hmdb-4} & 85.70\% & Yes \\ \hline
Ours + ResNext101 BERT \cite{hmdb-5} & 84.53\% & No \\ \hline
SMART \cite{hmdb-6} & 84.36\% & No \\ \hline
UniFormer \cite{uniformer} & 85.36\% & No \\ \hline
ViVit \cite{video-vit} & 80.18\% & No \\ \hline
\textbf{Ours} & \textbf{90.05\%} & No \\ \hline
\end{tabular}
%}

\end{table}

\hspace{1em}
The table \ref{tab:comp-us-vs-others-hmdb51} represents a comparison between our architecture and the state-of-the-art architectures on HMDB51, we added the results of Uniformer and Vivit since they were not tested on HMDB51. Our architecture improved the previous state of the art by $2.49\%$ and achieved a new SOTA result of $90.05\%$.

\begin{table}[!h]
\caption{Accuracy comparison on UCF101}
\label{tab:comp-us-vs-others-ucf101}
\centering
%\resizebox{0.73\width}{!}{%
\begin{tabular}{|l|c|c|}
\hline
\textbf{Architecture} & \textbf{Accuracy} & \textbf{Pre Training} \\ \hline
Localized Phase Features \cite{ucf-comp-1} & 99.21\% & No \\ \hline
SMART \cite{hmdb-6} & 98.64\% & No \\ \hline
OmniSource \cite{ucf-comp-3} & 98.60\% & Yes \\ \hline
PERF-Net \cite{ucf-comp-4} & 98.60\% & No \\ \hline
LGD-3D Two-stream \cite{ucf-comp-5} & 98.2\% & No \\ \hline
UniFormer \cite{uniformer} & - & No \\ \hline
ViVit \cite{video-vit} & - & No \\ \hline
\textbf{The proposed architecture} & \textbf{99.6\%} & No \\ \hline
\end{tabular}
\end{table}

\hspace{1em}
The table \ref{tab:comp-us-vs-others-ucf101} represents a comparison between our architecture and seven other state-of-the-art architectures tested on UCF101. The results of Uniformer and Vivit have not been included because they were not tested on UCF101. Our architecture improves the previous state of the art by 0.22\% and achieved a new SOTA result of $99.6\%$.

\begin{table}[!t]
\caption{Accuracy comparison on ETRI-Activity3D}
\label{tab:comp-us-vs-others-ETRI-Activity3D}
\centering
\begin{tabular}{|c|c|c|}
\hline
\textbf{Architecture} & \textbf{Data} & \textbf{Accuracy} \\ \hline
FSA-CNN \cite{etri-3d} & RGB & 90.1\% \\ \hline
FSA-CNN \cite{etri-3d} & RGB + S & 93.7\% \\ \hline
\textbf{Our Architecture} & \textbf{RGB} & \textbf{95.09\%} \\ \hline
\end{tabular}
\end{table}

The table \ref{tab:comp-us-vs-others-ETRI-Activity3D} represents a comparison between our architecture and the FSA-CNN \cite{etri-3d} architecture. FSA-CNN is a CNN architecture that takes as input videos (RGB) and/or skeleton data. It is based on a deep CNN network and an innovative approach to replace the activation functions.
%\hspace{1em}
Our model outperforms the FSA-CNN \cite{etri-3d} RGB model by more than $4.9\%$ and FSA-CNN RGB+S by $1.39\%$. 
Although the authors in \cite{etri-3d} claim to include spatiotemporal variation in action data, a CNN network that takes RGB video and skeleton data as input is not able to extract a complete spatiotemporal representation due to the nature of CNN and the complexity of the data (2 modalities of very different types). Our architecture outperforms FSA-CNN thanks to the proposed transformer that supports the extraction of spatiotemporal dependencies.

\section*{Conclusion and perspectives}

In this paper, we propose a novel sequential combination of 3D CNN convolution and a spatiotemporal transformer. Experiments show that the proposed architecture achieves new SOTA results with $90.05\%$, $99.6\%$, and $95.09\%$ on HMDB51, UCF101, and ETRI-Activity3D datasets respectively. In future work, we will investigate further schemes for combining CNN architectures with transformers. 

\bibliographystyle{IEEEtran}
\bibliography{MMSP2023/main}

% Generated by IEEEtran.bst, version: 1.12 (2007/01/11)
\begin{thebibliography}{10}
\providecommand{\url}[1]{#1}
\csname url@samestyle\endcsname
\providecommand{\newblock}{\relax}
\providecommand{\bibinfo}[2]{#2}
\providecommand{\BIBentrySTDinterwordspacing}{\spaceskip=0pt\relax}
\providecommand{\BIBentryALTinterwordstretchfactor}{4}
\providecommand{\BIBentryALTinterwordspacing}{\spaceskip=\fontdimen2\font plus
\BIBentryALTinterwordstretchfactor\fontdimen3\font minus \fontdimen4\font\relax}
\providecommand{\BIBforeignlanguage}[2]{{%
\expandafter\ifx\csname l@#1\endcsname\relax
\typeout{** WARNING: IEEEtran.bst: No hyphenation pattern has been}%
\typeout{** loaded for the language `#1'. Using the pattern for}%
\typeout{** the default language instead.}%
\else
\language=\csname l@#1\endcsname
\fi
#2}}
\providecommand{\BIBdecl}{\relax}
\BIBdecl

\bibitem{uniformer}
\BIBentryALTinterwordspacing
K.~Li, Y.~Wang, P.~Gao, G.~Song, Y.~Liu, H.~Li, and Y.~Qiao, ``Uniformer: Unified transformer for efficient spatiotemporal representation learning,'' \emph{CoRR}, vol. abs/2201.04676, 2022. [Online]. Available: \url{https://arxiv.org/abs/2201.04676}
\BIBentrySTDinterwordspacing

\bibitem{video-vit}
\BIBentryALTinterwordspacing
A.~Arnab, M.~Dehghani, G.~Heigold, C.~Sun, M.~Lucic, and C.~Schmid, ``Vivit: {A} video vision transformer,'' \emph{CoRR}, vol. abs/2103.15691, 2021. [Online]. Available: \url{https://arxiv.org/abs/2103.15691}
\BIBentrySTDinterwordspacing

\bibitem{dataset-hmdb51}
H.~Kuehne, H.~Jhuang, E.~Garrote, T.~Poggio, and T.~Serre, ``Hmdb: A large video database for human motion recognition,'' in \emph{2011 International Conference on Computer Vision}, 2011, pp. 2556--2563.

\bibitem{dataset-ucf-101}
\BIBentryALTinterwordspacing
K.~Soomro, A.~R. Zamir, and M.~Shah, ``{UCF101:} {A} dataset of 101 human actions classes from videos in the wild,'' \emph{CoRR}, vol. abs/1212.0402, 2012. [Online]. Available: \url{http://arxiv.org/abs/1212.0402}
\BIBentrySTDinterwordspacing

\bibitem{etri-3d}
\BIBentryALTinterwordspacing
J.~Jang, D.~Kim, C.~Park, M.~Jang, J.~Lee, and J.~Kim, ``Etri-activity3d: {A} large-scale {RGB-D} dataset for robots to recognize daily activities of the elderly,'' \emph{CoRR}, vol. abs/2003.01920, 2020. [Online]. Available: \url{https://arxiv.org/abs/2003.01920}
\BIBentrySTDinterwordspacing

\bibitem{vision-transformer}
\BIBentryALTinterwordspacing
A.~Dosovitskiy, L.~Beyer, A.~Kolesnikov, D.~Weissenborn, X.~Zhai, T.~Unterthiner, M.~Dehghani, M.~Minderer, G.~Heigold, S.~Gelly, J.~Uszkoreit, and N.~Houlsby, ``An image is worth 16x16 words: Transformers for image recognition at scale,'' \emph{CoRR}, vol. abs/2010.11929, 2020. [Online]. Available: \url{https://arxiv.org/abs/2010.11929}
\BIBentrySTDinterwordspacing

\bibitem{rw-cnn-1}
\BIBentryALTinterwordspacing
D.~Tran, H.~Wang, L.~Torresani, J.~Ray, Y.~LeCun, and M.~Paluri, ``A closer look at spatiotemporal convolutions for action recognition,'' \emph{CoRR}, vol. abs/1711.11248, 2017. [Online]. Available: \url{http://arxiv.org/abs/1711.11248}
\BIBentrySTDinterwordspacing

\bibitem{rw-cnn-2}
\BIBentryALTinterwordspacing
Z.~Qiu, T.~Yao, and T.~Mei, ``Learning spatio-temporal representation with pseudo-3d residual networks,'' \emph{CoRR}, vol. abs/1711.10305, 2017. [Online]. Available: \url{http://arxiv.org/abs/1711.10305}
\BIBentrySTDinterwordspacing

\bibitem{i3D-cnn}
\BIBentryALTinterwordspacing
J.~Carreira and A.~Zisserman, ``Quo vadis, action recognition? {A} new model and the kinetics dataset,'' \emph{CoRR}, vol. abs/1705.07750, 2017. [Online]. Available: \url{http://arxiv.org/abs/1705.07750}
\BIBentrySTDinterwordspacing

\bibitem{rw-cnn-3}
\BIBentryALTinterwordspacing
D.~Tran, H.~Wang, L.~Torresani, and M.~Feiszli, ``Video classification with channel-separated convolutional networks,'' \emph{CoRR}, vol. abs/1904.02811, 2019. [Online]. Available: \url{http://arxiv.org/abs/1904.02811}
\BIBentrySTDinterwordspacing

\bibitem{rw-cnn-3-1}
\BIBentryALTinterwordspacing
C.~Feichtenhofer, H.~Fan, J.~Malik, and K.~He, ``Slowfast networks for video recognition,'' \emph{CoRR}, vol. abs/1812.03982, 2018. [Online]. Available: \url{http://arxiv.org/abs/1812.03982}
\BIBentrySTDinterwordspacing

\bibitem{rw-cnn-4}
\BIBentryALTinterwordspacing
H.~Wang, D.~Tran, L.~Torresani, and M.~Feiszli, ``Video modeling with correlation networks,'' \emph{CoRR}, vol. abs/1906.03349, 2019. [Online]. Available: \url{http://arxiv.org/abs/1906.03349}
\BIBentrySTDinterwordspacing

\bibitem{rw-cnn-5}
X.~Li, Y.~Wang, Z.~Zhou, and Y.~Qiao, ``Smallbignet: Integrating core and contextual views for video classification,'' in \emph{Proceedings of the IEEE/CVF Conference on Computer Vision and Pattern Recognition}, 2020, pp. 1092--1101.

\bibitem{rw-cnn-6}
\BIBentryALTinterwordspacing
L.~Wang, Z.~Tong, B.~Ji, and G.~Wu, ``{TDN:} temporal difference networks for efficient action recognition,'' \emph{CoRR}, vol. abs/2012.10071, 2020. [Online]. Available: \url{https://arxiv.org/abs/2012.10071}
\BIBentrySTDinterwordspacing

\bibitem{rw-transformer-1}
\BIBentryALTinterwordspacing
G.~Bertasius, H.~Wang, and L.~Torresani, ``Is space-time attention all you need for video understanding?'' \emph{CoRR}, vol. abs/2102.05095, 2021. [Online]. Available: \url{https://arxiv.org/abs/2102.05095}
\BIBentrySTDinterwordspacing

\bibitem{rw-transformer-2}
\BIBentryALTinterwordspacing
D.~Neimark, O.~Bar, M.~Zohar, and D.~Asselmann, ``Video transformer network,'' \emph{CoRR}, vol. abs/2102.00719, 2021. [Online]. Available: \url{https://arxiv.org/abs/2102.00719}
\BIBentrySTDinterwordspacing

\bibitem{rw-transformer-3}
\BIBentryALTinterwordspacing
G.~Sharir, A.~Noy, and L.~Zelnik{-}Manor, ``An image is worth 16x16 words, what is a video worth?'' \emph{CoRR}, vol. abs/2103.13915, 2021. [Online]. Available: \url{https://arxiv.org/abs/2103.13915}
\BIBentrySTDinterwordspacing

\bibitem{rw-transformer-4}
\BIBentryALTinterwordspacing
X.~Zha, W.~Zhu, T.~Lv, S.~Yang, and J.~Liu, ``Shifted chunk transformer for spatio-temporal representational learning,'' \emph{CoRR}, vol. abs/2108.11575, 2021. [Online]. Available: \url{https://arxiv.org/abs/2108.11575}
\BIBentrySTDinterwordspacing

\bibitem{pmlr-v139-bertasius21a}
G.~Bertasius, H.~Wang, and L.~Torresani, ``Is space-time attention all you need for video understanding?'' in \emph{Proceedings of the 38th International Conference on Machine Learning}, ser. Proceedings of Machine Learning Research, M.~Meila and T.~Zhang, Eds., vol. 139.\hskip 1em plus 0.5em minus 0.4em\relax PMLR, 18--24 Jul 2021, pp. 813--824.

\bibitem{swin-transformer}
\BIBentryALTinterwordspacing
Z.~Liu, Y.~Lin, Y.~Cao, H.~Hu, Y.~Wei, Z.~Zhang, S.~Lin, and B.~Guo, ``Swin transformer: Hierarchical vision transformer using shifted windows,'' \emph{CoRR}, vol. abs/2103.14030, 2021. [Online]. Available: \url{https://arxiv.org/abs/2103.14030}
\BIBentrySTDinterwordspacing

\bibitem{hmdb-1}
\BIBentryALTinterwordspacing
L.~Wang and P.~Koniusz, ``Hallucinating statistical moment and subspace descriptors from object and saliency detectors for action recognition,'' \emph{CoRR}, vol. abs/2001.04627, 2020. [Online]. Available: \url{https://arxiv.org/abs/2001.04627}
\BIBentrySTDinterwordspacing

\bibitem{hmdb-2}
\BIBentryALTinterwordspacing
P.~Koniusz, L.~Wang, and K.~Sun, ``High-order tensor pooling with attention for action recognition,'' \emph{CoRR}, vol. abs/2110.05216, 2021. [Online]. Available: \url{https://arxiv.org/abs/2110.05216}
\BIBentrySTDinterwordspacing

\bibitem{hmdb-3}
\BIBentryALTinterwordspacing
P.~Koniusz, L.~Wang, and A.~Cherian, ``Tensor representations for action recognition,'' \emph{CoRR}, vol. abs/2012.14371, 2020. [Online]. Available: \url{https://arxiv.org/abs/2012.14371}
\BIBentrySTDinterwordspacing

\bibitem{hmdb-4}
\BIBentryALTinterwordspacing
P.~Koniusz, L.~Wang, and K.~Sun, ``High-order tensor pooling with attention for action recognition,'' \emph{CoRR}, vol. abs/2110.05216, 2021. [Online]. Available: \url{https://arxiv.org/abs/2110.05216}
\BIBentrySTDinterwordspacing

\bibitem{hmdb-5}
\BIBentryALTinterwordspacing
A.~Shah, S.~K. Mishra, A.~Bansal, J.~Chen, R.~Chellappa, and A.~Shrivastava, ``Pose and joint-aware action recognition,'' \emph{CoRR}, vol. abs/2010.08164, 2020. [Online]. Available: \url{https://arxiv.org/abs/2010.08164}
\BIBentrySTDinterwordspacing

\bibitem{hmdb-6}
\BIBentryALTinterwordspacing
S.~N. Gowda, M.~Rohrbach, and L.~Sevilla{-}Lara, ``{SMART} frame selection for action recognition,'' \emph{CoRR}, vol. abs/2012.10671, 2020. [Online]. Available: \url{https://arxiv.org/abs/2012.10671}
\BIBentrySTDinterwordspacing

\bibitem{ucf-comp-1}
\BIBentryALTinterwordspacing
S.~M. Hejazi and C.~Abhayaratne, ``Handcrafted localized phase features for human action recognition,'' \emph{Image and Vision Computing}, vol. 123, p. 104465, 2022. [Online]. Available: \url{https://www.sciencedirect.com/science/article/pii/S0262885622000944}
\BIBentrySTDinterwordspacing

\bibitem{ucf-comp-3}
\BIBentryALTinterwordspacing
H.~Duan, Y.~Zhao, Y.~Xiong, W.~Liu, and D.~Lin, ``Omni-sourced webly-supervised learning for video recognition,'' \emph{CoRR}, vol. abs/2003.13042, 2020. [Online]. Available: \url{https://arxiv.org/abs/2003.13042}
\BIBentrySTDinterwordspacing

\bibitem{ucf-comp-4}
\BIBentryALTinterwordspacing
Y.~Li, Z.~Lu, X.~Xiong, and J.~Huang, ``Perf-net: Pose empowered rgb-flow net,'' \emph{CoRR}, vol. abs/2009.13087, 2020. [Online]. Available: \url{https://arxiv.org/abs/2009.13087}
\BIBentrySTDinterwordspacing

\bibitem{ucf-comp-5}
\BIBentryALTinterwordspacing
Z.~Qiu, T.~Yao, C.~Ngo, X.~Tian, and T.~Mei, ``Learning spatio-temporal representation with local and global diffusion,'' \emph{CoRR}, vol. abs/1906.05571, 2019. [Online]. Available: \url{http://arxiv.org/abs/1906.05571}
\BIBentrySTDinterwordspacing

\end{thebibliography}
\end{document}